\documentclass[10pt,twocolumn,letterpaper]{article}
\usepackage{cvpr}         
\usepackage{amsmath, amssymb}
\usepackage{algorithm}
\usepackage[noend]{algpseudocode}
\usepackage{amssymb}
\usepackage{booktabs}
\usepackage{multirow}

\usepackage{xspace}

\definecolor{cvprblue}{rgb}{0.21,0.49,0.74}
\usepackage[pagebackref,breaklinks,colorlinks,allcolors=cvprblue]{hyperref}

\newcommand{\method}{MeshFlow\xspace}

\makeatletter
\renewcommand{\paragraph}{%
  \@startsection{paragraph}{4}%
  {\z@}{-0.5em}{-0.5em}%
  {\normalfont\normalsize\bfseries}%
}
\makeatother

\def\paperID{14612} 
\def\confName{CVPR}
\def\confYear{2026}

\title{\method: Efficient Artistic Mesh Generation via MeshVAE and Flow-based Diffusion Transformer}

\author{
Weiyu Li\textsuperscript{1,2} \quad 
Antoine Toisoul\textsuperscript{1} \quad 
Tom Monnier\textsuperscript{1} \quad 
Roman Shapovalov\textsuperscript{1} \quad 
Rakesh Ranjan\textsuperscript{1} \\
Ping Tan\textsuperscript{2$\dagger$} \quad 
Andrea Vedaldi\textsuperscript{1$\dagger$} \\
\small \textsuperscript{1}Meta AI \quad \textsuperscript{2}Hong Kong University of Science and Technology \\
\small \textsuperscript{$\dagger$}Corresponding authors
}

\begin{document}

\twocolumn[{%
\renewcommand\twocolumn[1][]{#1}%
\maketitle
\begin{center}
\centering
\includegraphics[width=1.\textwidth]{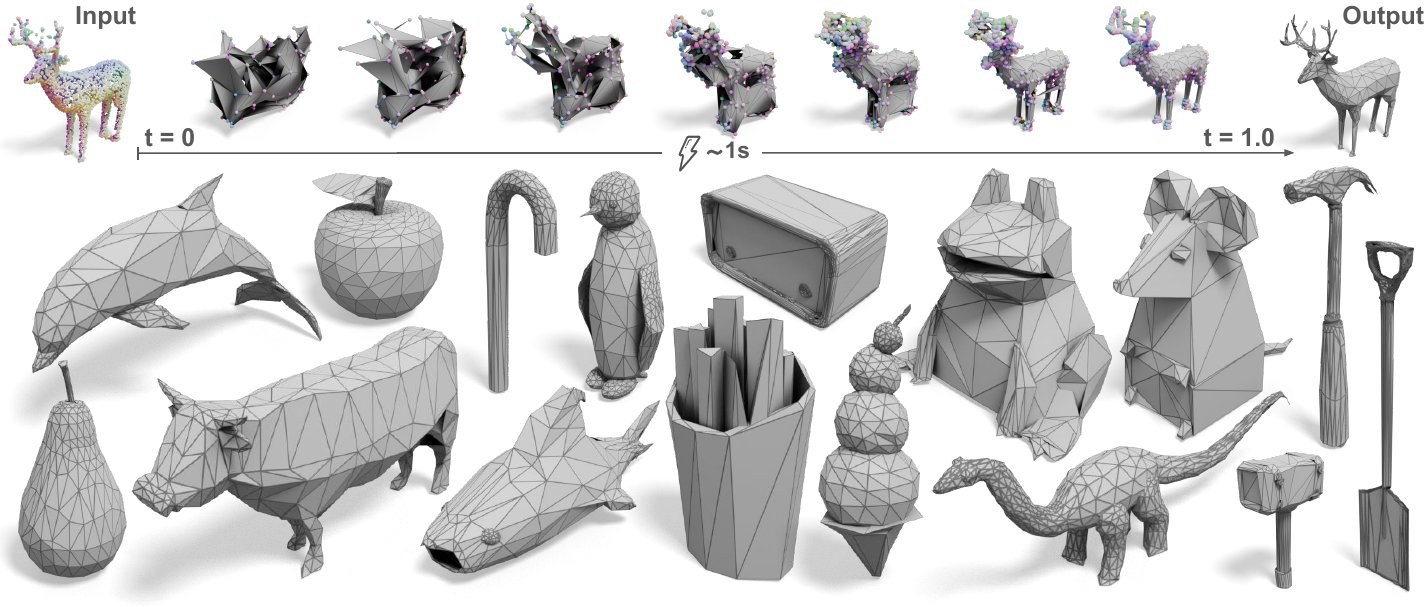}
\vspace{-3em}
\captionof{figure}{
Compared to prior, autoregressive, mesh-generation models, our proposed \method generates high-fidelity 3D meshes in approximately $1$ second. At its core is a novel MeshVAE that compresses discrete mesh data---including vertices, vertex normals, and connectivity---into a compact, continuous latent space, and then uses rectified flow for efficient generation.
}%
\label{fig:teaser}
\end{center}%
}]
 
\begin{abstract}
We present \method, a new method for generating artist-like 3D meshes.
Current mesh generators often adopt Auto-Regressive (AR) next-token prediction, a natural choice given the discrete nature of mesh topology.
However, AR methods scale poorly because the inference cost is quadratic in mesh size.
They also require discretizing the vertex coordinates, which introduces quantization errors.
To address these challenges, we introduce a Variational Autoencoder (VAE) that, supervised with a contrastive loss, represents both continuous vertex positions and discrete connectivity in a continuous latent space.
This latent space is significantly more compact than prior token-based mesh representations.
We then build a 3D generator based on a Rectified Flow transformer, generating all mesh vertices and edges in parallel.
Our model generates meshes $18\times$ faster than the fastest AR generator while also achieving excellent accuracy across standard mesh-generation metrics.
\end{abstract}
    
\vspace{-2em}
\section{Introduction}%
\label{sec:intro}

Three-dimensional meshes are the standard representation of 3D shapes in AR/VR/XR, video games, visual effects (VFX), and many other applications.
Yet, crafting high-quality, artist-grade 3D meshes requires specialized tools, significant expertise, and time.
While progress in generative modeling has dramatically reduced barriers to content creation in language, images, and other modalities, giving hope that this technology could also be applied to the production of artist-grade 3D meshes, achieving high-quality results in mesh generation remains an open problem.

A challenge is that meshes have continuous vertices but discrete topology, which is difficult to generate directly.
For this reason, many 3D generators~\cite{zhang2023shape2vecset,zhang2024clay,xiang2024structured,li2025triposg,hunyuan3d25hunyuan3d,li2025step1x} have focused on predicting implicit shape representations such as Signed Distance Function (SDF), extracting meshes from those in post-processing, usually by means of Marching Cubes~\cite{marchingcubes}.
However, implicit representations can impose unwanted constraints, such as requiring 3D shapes to be watertight.
Most importantly, the structure of meshes extracted in this manner is hardly satisfactory.
Such meshes tend to smooth out sharp edges or contain far too many vertices and faces, making them inefficient for real-time applications and difficult for artists to manipulate and edit.

The alternative is to generate better 3D meshes \emph{directly}.
Because the mesh topology (i.e., the way vertices are connected into edges and faces) is discrete, many authors~\cite{nash2020polygen,siddiqui2023meshgpt,chen2024meshanything,weng2024bpt} have noted an analogy with Large Language Models (LLMs), as text is also discrete.
Inspired by LLMs, they discretize and tokenize the vertices and faces of the mesh, and then generate them in an Auto-Regressive (AR) manner.
However, this paradigm has several drawbacks.
First, the most basic mesh tokenizers associate a token with each coordinate in the mesh.
Since each triangular face has three vertices, and each vertex has three coordinates, a mesh with $n_\mathbf{f}$ faces is represented by at least $9n_\mathbf{f}$ tokens~\cite{chen2024meshxl, chen24meshanything,hao2024meshtron}.
Although less na\"{\i}ve tokenizers~\cite{song2025meshsilksong} can reduce this number by $78\%$, AR generation still limits the model's ability to output large and detailed meshes in a reasonable time as, ultimately, the inference cost grows quadratically with the mesh size.
Second, AR models can terminate sequences prematurely, resulting in incomplete meshes.
Finally, they discretize the vertex coordinates, typically using only $128$ levels~\cite{weng2024bpt,chen2024meshanything,song2025meshsilksong,kim2025fastmesh,wang2025nautilus} for efficiency.
This introduces quantization errors, occasionally causing vertices to collapse and faces to overlap.

In this paper, we thus seek a much more efficient, non-AR approach to mesh generation.
Inspired by latent diffusion models for image and video generation, which can efficiently output millions of image pixels in parallel, we develop a \emph{compact and continuous latent space for meshes}, which we call \method.
Its design starts by noting that a watertight mesh can be represented as a collection of vertices, the edges connecting them, and a circular ordering of edges around each vertex, from which the mesh faces and topology can be inferred~\cite{muller78finding}.
The challenge, then, is to encode the edges as a pair of vertex indices, which are discrete.
Similarly to~\cite{spacemesh2024}, we sidestep this issue by representing edge connections in a continuous manner, by associating a feature vector to each vertex, and assuming that two vertices are connected by an edge if, and only if, the distance between their feature vectors is below a certain threshold.
We use contrastive learning to learn these vectors.
We encode the ordering of the edges around each vertex by predicting an additional attribute: vertex normals, which define a clockwise rotation around each vertex.

\begin{figure}[t!]
\centering
\includegraphics[width=\linewidth]{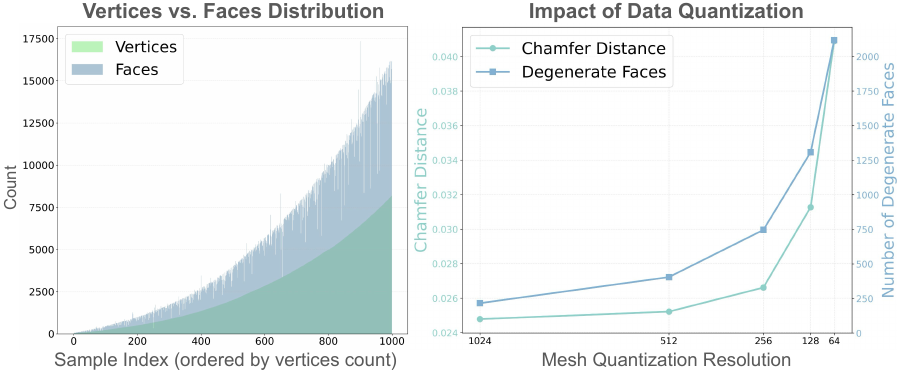}
\caption{
Data Statistics Visualization.
(Left) Distribution of vertices and faces, showing that the face count is roughly double the vertex count.
(Right) Impact of reduced quantization resolution, illustrating that lower resolution leads to increased geometric errors and face collapse.
}%
\label{fig:data_statistics}
\vspace{-5mm}
\end{figure}

As a result, the mesh is encoded by \emph{assigning a continuous latent vector to each vertex}, comprising the vertex position, normal, and edge embedding.
Hence, a mesh with $n_\mathbf{v}$ vertices is represented by $n_\mathbf{v}$ such vectors.
As shown in \cref{fig:data_statistics} (left), because there are typically two to three times more faces than vertices, this encoding uses fewer components than the face-oriented tokenizers employed in prior work.
In addition, our representation is continuous, which is the key to efficient generation by means of latent denoising diffusion, as we discuss next.

To apply latent diffusion, we learn a Variational Autoencoder (VAE), dubbed \emph{MeshVAE}, to map our continuous mesh representation to an even more compact latent space.
Inspired by~\cite{zhang233dshape2vecset:}, our encoder packs mesh information in a subset of the vertices, thus downsampling the representation.
Our experiments demonstrate that, even by retaining only $n_\mathbf{v} / 4$ latent vectors, the \emph{MeshVAE} decoder can still accurately reconstruct the mesh.
Our representation has 72$\times$ fewer token vectors than na\"{\i}ve tokenizers and 16$\times$ fewer than the most efficient one.
In addition to achieving much better compression, our representation does not discretize the vertices of the mesh, thus avoiding quantization errors.
Using this encoding, we then train a Rectified Flow transformer for generating 3D meshes from point clouds.
Compared to AR generators, the resulting generator is much faster, and its inference cost grows only linearly with mesh size rather than quadratically.

To summarize, we make the following contributions:
\begin{itemize}
\item We introduce MeshVAE, a novel autoencoder that compresses 3D meshes into a compact, discretization-free latent space, achieving excellent reconstruction quality with a significantly smaller latent size than existing mesh tokenizers.
\item We show that a Rectified Flow generator that uses the proposed VAE achieves SoTA performance on standard artist-like mesh generation benchmarks, along with a significant $18\times$ speedup in inference time.
\item We show applications to point cloud and image-conditioned mesh generation.
\end{itemize}

\section{Related Work}%
\label{sec:related}

We discuss prior works related to the generation of 3D meshes.
We first provide a brief overview of relevant literature on mesh representations, followed by a discussion of recent works focused on artistic mesh generation.

\paragraph{Indirect Mesh Generation.}

Rather than generating meshes directly, many works~\cite{mildenhall2020nerf,yariv2021volume,wang2021neus,Liu23NeUDF,zhang2023shape2vecset} use differentiable shape representations~\cite{mildenhall2020nerf,yariv2021volume,wang2021neus,Liu23NeUDF,zhang2023shape2vecset} like distance fields, and then use iso-surface extraction algorithms~\cite{marchingcubes,shen2023flexicubes,chen2022ndc} to extract the mesh. %
Although these generators produce compelling \emph{shapes}, the resulting \emph{meshes} are far from being as polished and well-structured as those created by artists.
Generative models for Computer-Aided Design (CAD)~\cite{xu2024brepgen,CLRWire25} focus on generating parameterized curves and have achieved significant progress in producing topologies with sharp, well-defined edges.
However, they are inherently constrained to linear structures and struggle to represent complex, free-form geometries.

\paragraph{Autoregressive Mesh Generation.}

Most mesh generators rely on auto-regressive (AR) architectures inspired by language models, which are well-suited to the discrete nature of meshes.
PolyGen~\cite{nash2020polygen} and MeshGPT~\cite{siddiqui2023meshgpt} pioneered this approach, while MeshXL~\cite{chen2024meshxl} and MeshAnything~\cite{chen2024meshanything} added support for conditional generation based on input point clouds.
One limitation is the inefficient mesh tokenizer, which represents each vertex coordinate of each face separately, for a total of $9n_\mathbf{f}$ tokens.
Meshtron~\cite{hao2024meshtron} mitigates this by using a hierarchical Hourglass Transformer~\cite{crowson24scalable} to efficiently process a large number of tokens.
EdgeRunner~\cite{tang2024edgerunner} combines an Auto-regressive Auto-Encoder (ArAE) with latent diffusion, but the discrete ArAE fails to represent the mesh accurately or compactly.
Other works~\cite{weng2024bpt,lionar2025treemeshgpt,song2025meshsilksong} propose more efficient tokenizers.
In particular, BPT~\cite{weng2024bpt} introduces block-wise indexing and patchified aggregation.
TreeMeshGPT~\cite{lionar2025treemeshgpt} uses a sequencing strategy based on a dynamically growing tree structure, and MeshSilksong~\cite{song2025meshsilksong} proposes a hierarchical tokenizer, with both yielding a $78\%$ smaller encoding than the na\"{\i}ve approach.
FastMesh~\cite{kim2025fastmesh} generates vertices using an AR model, only requiring $n_\mathbf{v}$ tokens, but predicts faces deterministically, even though faces are not uniquely determined by the vertices in general.

Overall, AR-based methods scale poorly with the size of the generated mesh and quantize vertices, losing precision and risking vertex collapse.
Our approach uses only $n_\mathbf{v} / 4$ tokens, avoids quantization, and supports generation via denoising diffusion, which is much more efficient.

\paragraph{Differentiable Mesh Representations.}

Other works have also developed differentiable mesh representations that can be fit to shape data using gradient descent.
For instance, DMesh~\cite{son2024dmesh} introduces a differentiable formulation for the probability that a face exists, and DMesh++~\cite{son2024dmeshepp} accelerates it by introducing the Minimum-Ball algorithm.
However, both approaches are limited to meshes that can be obtained by Delaunay triangulation.

\paragraph{Non-autoregressive Mesh Generation.}

Similar to our method, other works have developed continuous mesh representations that can be processed by non-autoregressive models.
Ploydiff~\cite{alliegro2023polydiffgenerating3dpolygonal} performs on quantized triangle soups.
MeshCraft~\cite{he2025meshcraft} computes such a representation using a Graph Convolutional Network (GCN) autoencoder, and uses a Diffusion Transformer (DiT)~\cite{peebles23scalable} to generate the mesh.
The GCN acts as a bottleneck, hindering generation of meshes with more than $1{,}536$ faces.
PDT~\cite{pdt} also uses a DiT to generate the mesh vertex locations, but relies on traditional Quadric Error Metrics (QEM) simplification~\cite{qem}, so the resulting topology fails to match the data distribution.
LATO~\cite{zhao2026lato} proposes a topology-preserving latent representation for mesh diffusion.
These methods also still require discretization of mesh vertices.
SpaceMesh~\cite{spacemesh2024} introduces a continuous representation of topology which, similar to our method, is based on comparing per-vertex embeddings.
However, their face predictor does not model the stochasticity of topology.
Furthermore, they propose a relatively complex mechanism to encode the winding order of the edges around each vertex, whereas we simply predict the vertex normals to support this.
Our latent space is significantly more compact, which allows us to generate more complex meshes.
We also generate both vertex positions and connectivity in parallel using denoising diffusion.

\section{Method}%
\label{sec:method}

\begin{figure*}[t!]
\centering
\includegraphics[width=\linewidth]{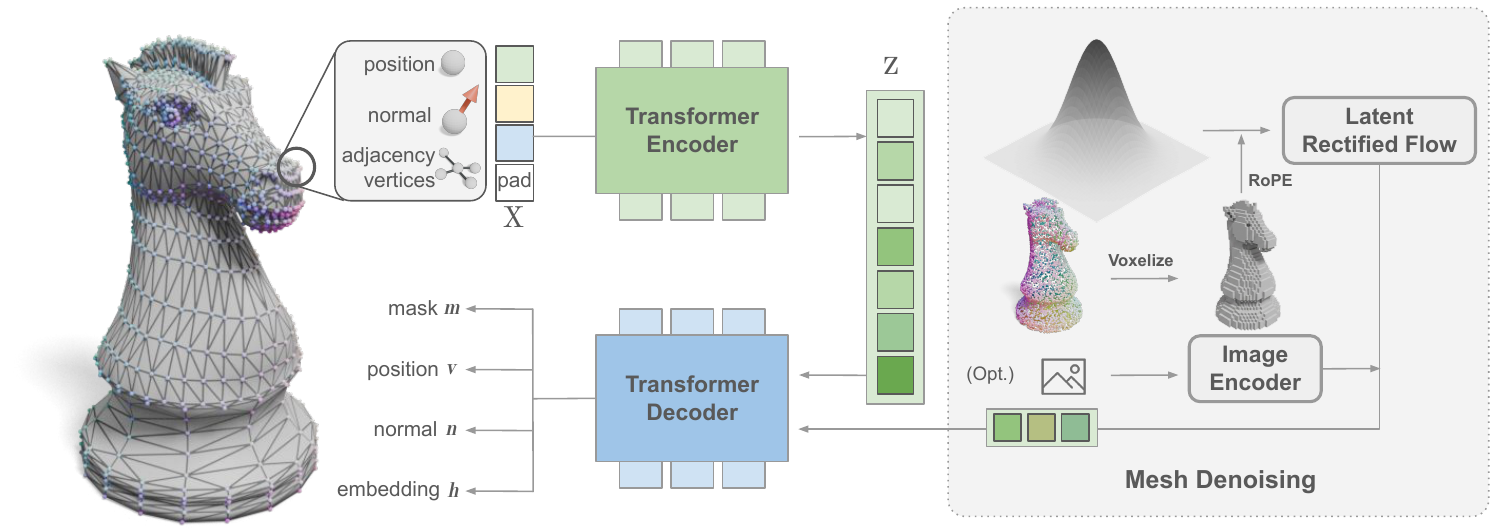}
\caption{
Overview of our method.
We first propose MeshVAE, which compresses vertices, vertex normals, and discrete adjacency relationships of a mesh into a continuous latent space.
This is supervised by the ground-truth vertices and vertex normals, coupled with a contrastive learning approach applied to vertex adjacency.
We then employ latent Rectified Flow based on the proposed representation, and finally pass the result through the Mesh Decoder to obtain a mesh.
}%
\label{fig:overview}
\end{figure*}

We introduce \method, a novel framework for efficient artist-like mesh generation.
Instead of adopting an AR model, which scales poorly and requires quantizing vertices, we propose a new auto-encoder that computes a continuous latent representation of meshes (\cref{sec:mesh_representation}), and use the latter in a Rectified Flow~\cite{liu23flow} Diffusion Transformer (DiT)~\cite{peebles23scalable} for efficient generation (\cref{sec:mesh_generation}).

\subsection{Mesh Representation}%
\label{sec:mesh_representation}

\newcommand{\adjacency}{\mathcal{A}} 
\newcommand{\decoder}{\mathcal{D}}
\newcommand{\edge}{\boldsymbol{e}}
\newcommand{\edges}{E}
\newcommand{\embedding}{\boldsymbol{h}}
\newcommand{\encoder}{\mathcal{E}}
\newcommand{\face}{\boldsymbol{f}}
\newcommand{\faces}{F}
\newcommand{\latent}{\boldsymbol{z}}
\newcommand{\mask}{\boldsymbol{m}}
\newcommand{\mesh}{\mathcal{M}}
\newcommand{\neighbours}{\mathcal{N}} 
\newcommand{\normal}{\boldsymbol{n}}
\newcommand{\order}{\pi}
\newcommand{\real}{\mathbb{R}}
\newcommand{\vertex}{\boldsymbol{v}}
\newcommand{\vertices}{V}

A \emph{triangular mesh} $\mesh$ can be represented as a list of vertices
$
\vertices = (\vertex_1,\dots,\vertex_N)
$,
where each vertex $\vertex_i \in \real^3$ is a 3D point, along with a set of faces $\faces$, where each face $\face \in \faces$ is a set of three integers 
$
\face = \{f_1, f_2, f_3\}
$
taken from the set $\{1,\dots,N\}$ which indexes the vertices in $\vertices$ that form the corners of triangular faces.

Instead of encoding faces directly, we propose to encode their edges
$\edge \in \edges$.
Formally, the edges of the mesh $\mesh$ are all unordered pairs of face vertices:
$
\edges = \{
\{e_1,e_2\}: 
e_1 \not= e_2 \wedge \exists \face \in \faces :  \{e_1,e_2\}\subset \face
\}
$.
If we exclude meshes with triangular edge loops that do not bound a triangle, the faces $\faces$ can be recovered from the edges $\edges$ by looking for loops of three edges.
I.e., the faces $\faces$ are all unordered triplets  of indices supported by edges:
\begin{multline}\label{eq:back_to_faces}
\faces = \{ \{f_1,f_2,f_3\}: \{f_1,f_2\},~\{f_2,f_3\},~\{f_3,f_1\}  \in \edges \}.
\end{multline}
Hence, with negligible loss of generality, the mesh can be represented by the vertices $\vertices$ and the edges $\edges$.
In turn, edges are given by the \emph{adjacency matrix}
$
\mathcal{A} \in \{0, 1\}^{N \times N}
$,
where $\mathcal{A}_{ij} = 1$ iff there is an edge $(i,j)\in \edges$ between vertices $i$ and $j$.
However, the adjacency matrix is discrete and non-differentiable.
SpaceMesh~\cite{shen25spacemesh:} suggests instead assigning to each vertex an \emph{edge embedding} $\embedding_i \in \real^D$.
The existence of an edge between vertices $i$ and $j$ is then implicitly determined by computing the distance $d(\embedding_i, \embedding_j)$ between embeddings and applying a threshold $\tau$, so that the adjacency matrix is recovered as
\begin{equation}
\label{eq:mesh_continuous}
\mathcal{A}_{ij} = \mathbb{I} [d(\embedding_i, \embedding_j) \leq \tau],
\end{equation}
where $\mathbb{I}[\cdot]$ is the indicator function.
The advantage of this approach is that it expresses the discrete adjacency relations in terms of continuous (and learnable) vertex embeddings.

So far, we have considered faces as orderless triplets of indices, which, geometrically, means that the faces are unoriented.
Usually, however, we are interested in orienting the faces to define an inside and an outside of the object.
This can be expressed by ordering the vertices of each face, so that they are triplets $(f_1,f_2,f_3)$ instead of sets.
Instead of encoding the order directly, however, we simply assign an \emph{outward normal vector} $\normal_i \in \mathbb{S}^2$ at each vertex.
Then, the normal of the face is obtained by averaging the normals of its vertices, and the face is oriented accordingly.

We have thus represented a mesh $\mesh$ as a triplet
$
\mesh = ( \vertex, \normal, \embedding )
$
where $(\vertex_i, \normal_i, \embedding_i)$ are the position, normal, and edge embedding associated to vertex $i=1,\dots,N$.
Given  $\embedding$, we can recover the adjacency matrix $\mathcal{A}$,
the edges $\edges$ by using \cref{eq:mesh_continuous},
the faces $\faces$ by using \cref{eq:back_to_faces},
and the face orientations by using $\normal$.

\paragraph{Comparison to SpaceMesh.}

We represent edges in a similar manner to SpaceMesh~\cite{shen25spacemesh:}, but our assumptions and the way we recover the topology of the mesh differ.
SpaceMesh considers meshes more general than triangular ones.
To do so, they use the \emph{half edge} representation, which in turn assumes that the mesh is manifold and watertight.
Many artist-like meshes do not satisfy this property, and we do not require it.
To encode the half edges, they need to represent the ordering of the edges around each vertex.
While this could presumably be inferred from the normal $\normal_i$, they instead introduce further continuous per-vertex embeddings
$
(
\boldsymbol{y}_i^\text{root},
\boldsymbol{y}_i^\text{prev},
\boldsymbol{y}_i^\text{next}
)
$
to do so, which we avoid.

\subsection{MeshVAE}%
\label{sec:contrastive_vae_training}

We now describe the design of MeshVAE\@.
The goal is to encode a mesh into a continuous latent vector $\latent$, and then to decode it back to reconstruct it. This VAE is `translational', in the sense that the input and output representations of the mesh differ (but are equivalent).
Specifically, the encoder $\latent = \encoder(\vertex, \normal, \adjacency)$ takes as input the vertices $\vertex$, their normals $\normal$, and the neighbor vertices extracted from the adjacency matrix $\adjacency$.
The decoder
$
(
  \hat{\vertex},
  \hat{\normal},
  \hat{\embedding},
  \hat{\mask}
)
=
\decoder(\latent)
$
instead outputs the vertices, normals, and edge embeddings (from which the topology can be recovered).
Furthermore, these are padded to a fixed maximum number of vertices $N$, and the decoder also outputs a mask $\mask \in [0,1]^N$ that indicates which vertices are valid. 
The padded vertices and normals are sampled randomly from the mesh, and the adjacency information is propagated from nearest vertices.
Note that capturing the topology via $\adjacency$, which is discrete, is not a problem as it is used as an input to the encoder; in the decoder the topology is captured by $\embedding$, which is continuos.

\begin{figure}[t!]
\centering
\includegraphics[width=\linewidth]{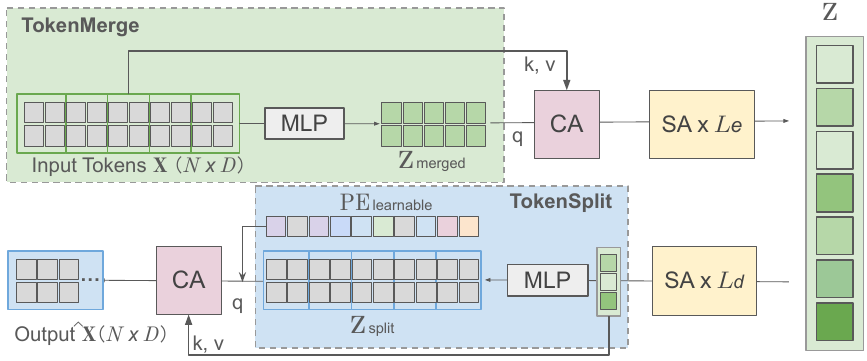}
\caption{
Detailed structure of our MeshVAE\@.
We found that using a simple TokenMerge and TokenSplit strategies for downsampling and upsampling enables more effective preservation of the original information, thereby achieving better reconstruction performance.
}%
\label{fig:detailed_vae}
\vspace{-5mm}
\end{figure}

\paragraph{Encoder Architecture.}

The encoder $\encoder$ first encodes the vertex and normal coordinates
using the Fourier Positional Encoding operator $\operatorname{PE}$.
We facilitate the encoder by associating to each vertex its normal and the other vertices that surround it.
Let $\neighbours_i \subset \{1,\dots,N\}$ be the indices of the neighbors of vertex $i$ according to the adjacency matrix $\adjacency$.
The feature vector is given by:
$
\boldsymbol{x}_i = \operatorname{Concat}
\left(
   \operatorname{PE}(\vertex_i),
   \operatorname{PE}(\normal_i),
   \operatorname{Concat}_{j \in \neighbours_i}(\vertex_j)
\right).
$
Note that the dimension of these vectors depends on the number of neighbors of the vertex, which can vary.
We therefore pad them to a fixed size $D$, assuming a maximum number of neighbors for all vertices.
We then concatenate $\boldsymbol{x}_i$ in an $N \times D$ matrix $\mathbf{X}$ of tokens, where $D$ is the dimensionality of each token vector $\boldsymbol{x}_i$.
As shown in \cref{fig:detailed_vae}, we then merge some of the $N$ tokens down to $n < N$ using a $\operatorname{MergeToken}$ operator inspired by the pixel-shuffle technique of InternVL~\cite{chen2024interlvl}.
This is followed by a Multi-Layer Perceptron (MLP) to obtain the encoding
$
\latent_\text{merged} = \operatorname{MLP}(\operatorname{MergeToken}(\mathbf{X})).
$
Then, inspired by VecSet~\cite{zhang233dshape2vecset:}, we let these reduced tokens attend to the full set of input tokens $\mathbf{X}$ using a Cross-Attention (CA) layer, so as to capture the details of the input mesh.
This is further passed through $L_e$ Self-Attention layers, which we denote as $\operatorname{SA}^{(L_e)}$, to obtain the final latent representation:
$$
\latent = \operatorname{SA}^{(L_e)}(
    \operatorname{CA}(\latent_{\text{merged}}, \mathbf{X})
  ).
$$

\paragraph{Decoder Architecture.}

The decoder $\decoder$ has an architecture symmetric to the encoder, aiming to reconstruct the mesh components $\hat{\vertex}$, $\hat{\normal}$, $\hat{\embedding}$, and $\hat{\mask}$ from the latent code $\latent$.
First, a $\operatorname{SplitToken}$ operator maps $n$ latent tokens $\latent$ back to $N$, obtaining
$
\latent_\text{split} = \operatorname{MLP}(\operatorname{SplitToken}(\latent)).
$
Then $\latent_\text{split}$ attends $\latent$ processed through $L_d$ Self-Attention layers via cross attention along with a learnable Positional Embedding to output the features $\hat{\mathbf{X}}$:
$$
\hat{\mathbf{X}} = \operatorname{CA}(\latent_\text{split} + \operatorname{PE}_\text{learnable}, \operatorname{SA}^{(L_d)}(\latent)).
$$
This is interpreted as the concatenation of $N$ vectors
$
\hat{\boldsymbol{x}}_i = (
  \hat{\vertex}_i,
  \hat{\normal}_i,
  \hat{\embedding}_i,
  \hat{\mask}_i
)
$
containing the desired information.

\begin{figure}[t!]
\centering
\includegraphics[width=\linewidth]{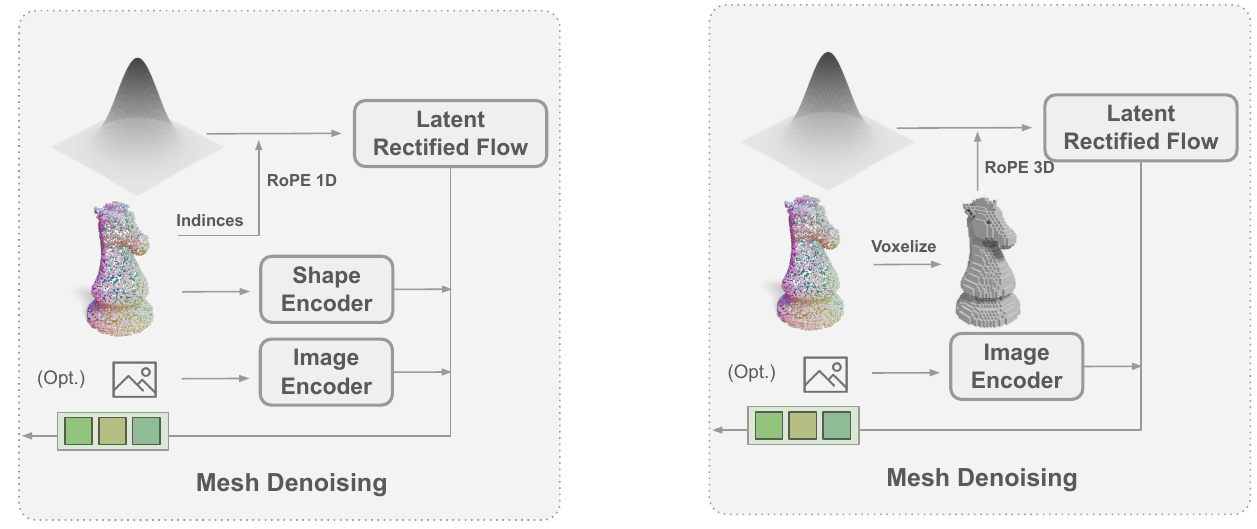}
\caption{
Two alternative strategies to inject shape information into DiT.
}%
\label{fig:different_shape_injection}
\vspace{-5mm}
\end{figure}

\paragraph{Training Objective.}

The VAE is trained to minimize the sum of reconstruction loss $\mathcal{L}_{\text{rec}}$ and KL regularization loss $\mathcal{L}_{\text{kl}}$, which forces the prior distribution of $\latent$ to match standard Gaussian distribution.
The overall reconstruction loss $\mathcal{L}_{\text{rec}}$ is a weighted sum of four distinct components, corresponding to the predicted vertices $\hat{\mathbf{v}}$, normals $\hat{\mathbf{n}}$, adjacency embeddings $\hat{\embedding}$ and vertex mask $\hat{\mathbf{m}}$:
\begin{equation}
  \mathcal{L}_{\text{rec}} = \mathcal{L}_{\text{mask}} + \mathcal{L}_{\mathbf{v}} + \mathcal{L}_{\mathbf{n}} + \mathcal{L}_{\text{adj}} + \lambda_{kl}\mathcal{L}_{kl},
\end{equation}
where we use Binary Cross-Entropy (BCE) on the predicted vertex mask $\mathcal{L}_{\text{mask}} = \operatorname{BCE}(\hat{\mathbf{m}}, \mathbf{m})$, and Mean Squared Error (MSE) is averaged over only valid vertices and normals.
A contrastive loss defined over the distances $d(\hat{\mathbf{e}}_i, \hat{\mathbf{e}}_j)$ between all pairs of predicted adjacency embeddings $\hat{\mathbf{e}}$ and a threshold $\tau$ for positive and negative edges:
\begin{equation}
\begin{gathered}
  \mathcal{L}_{\text{adj}} = \mathcal{L}_{\text{pos}} + \lambda_{\text{neg}} \cdot R_{\text{neg/pos}} \cdot \mathcal{L}_{\text{neg}}, \\
  \mathcal{L}_{\text{pos}} = -\frac{1}{|E|} \sum_{(i,j) \in E} \log\left(\sigma(d(\hat{\mathbf{e}}_i, \hat{\mathbf{e}}_j) - \tau)\right), \\
  \mathcal{L}_{\text{neg}} = -\frac{1}{|\neg E|} \sum_{(i,j) \notin E} \log\left(\sigma(\tau - d(\hat{\mathbf{e}}_i, \hat{\mathbf{e}}_j))\right),
\end{gathered}
\end{equation}
where $E$ and $\neg E$ are the sets of existing and non-existing edges within each mesh, respectively.
$R_{\text{neg/pos}} = |\neg E| / |E|$ is the ratio of negative to positive edges
and $\lambda_{\text{neg}}$ is used to balance these losses.
We use the Space-time Distance used by SpaceMesh~\cite{spacemesh2024} as the distance function $d$.

\subsection{Robust Face Recovery and Mesh Extraction}%
\label{sec:mesh_extraction}

We now discuss how the mesh is recovered from the representation $(\vertex, \normal, \embedding)$.
As noted in \cref{sec:mesh_representation}, we first obtain the set of valid edges based on the edge embeddings, and then check for triplets of edges that share three distinct vertices among them.
The resulting triplets are considered triangular faces.
The ordering of the vertices within this face is determined by calculating the face normal as the mean of the three vertex normals.

Furthermore, to repair defects caused by inaccurate diffusion predictions, we implement a post-processing step to repair boundaries.
A boundary edge is defined as an edge that belongs to only one triangular face.
We therefore detect boundary edges and then determine whether these boundary edges form a $k$-gon ring.
By default, for $k < 5$ we perform triangulation to convert the loop into multiple triangular faces, thereby enhancing the robustness of the generated results.
Please refer to the supplementary for more details.

\subsection{Mesh Generation with Flow Matching}%
\label{sec:mesh_generation}

We employ Rectified Flows (RF)~\cite{liu23flow} for fast and efficient mesh generation.
The advantage of RF lies in its straight-line ODE formulation, which avoids path crossing and minimizes discretization errors in the diffusion timesteps.
The model uses a linear forward process $\boldsymbol{x}(t)=(1-t)\boldsymbol{x}_0+t\boldsymbol{\epsilon}$ and approximates the backward vector field $\boldsymbol{v}(\boldsymbol{x},t) = \nabla_t\boldsymbol{x}$ using a neural network $\boldsymbol{v}_\theta$ trained with the Conditional Flow Matching (CFM)~\cite{lipman2023flow} objective:
\begin{equation}
  \mathcal{L}_{\text{CFM}}(\theta) = \mathbb{E}_{t, \boldsymbol{x}_0, \boldsymbol{\epsilon}} \left\| \boldsymbol{v}_\theta\Big(\mathrm{RoPE3D}(\boldsymbol{x}_t, \boldsymbol{c}_\textrm{vox}), \, t\Big) - (\boldsymbol{\epsilon}-\boldsymbol{x}_0) \right\|^2_2.
\end{equation}

For conditional generation, we adapt the Diffusion Transformer architecture~\cite{peebles2023scalable} conditioned on an input point cloud, which is first voxelized and then integrated with the noise, using a RoPE 3D positional encoding~\cite{su2023roformerenhancedtransformerrotary}. We also provide the number of vertices concatenated with the time embedding $t$ at each timestep to provide global conditioning. More details on the implementation are available in appendix. Similarly to the observations made in PDT~\cite{pdt} and MeshCraft~\cite{he2025meshcraft}, we found that effective denoising requires focusing on fine geometric details. We therefore adopt, during the final phase of training, the logit-normal sampling methodology introduced in SD3~\cite{esser2024scaling} and inference with timestep shifting $3.0$, to encourage the network to prioritize on the detailed geometry during generation.

\paragraph{The Choice of Shape Information Injection.}
We have two alternative strategies to inject shape information into our DiT as shown in Fig.~\ref{fig:different_shape_injection}. In our initial design, we uniformly sampled 32768 points for shape conditioning and then fed these points into a pre-trained shape encoder with an architecture analogous to Shape2VecSet~\cite{zhang2023shape2vecset}, yielding 2048 shape tokens. We then leveraged cross-attention (CA) to condition our DiT on these tokens and further added a vertex count condition in the tilmestep embedding. We found that although this paradigm facilitates straightforward inference, yet it requires prolonged training to ensure good model performance. Another approach encodes shape information using the XYZ coordinates of the ground truth vertices combined with Rotary Positional Embeddings (RoPE~\cite{su2023roformerenhancedtransformerrotary}) to give 3D spatial encodings to the noisy tokens during training. However, as the ground truth vertex distribution differs from the one of uniformly sampled point clouds, this approach introduces a domain gap between training and inference, resulting in outputs lacking artistic aesthetics. Nevertheless, we observe that this scheme enables much faster convergence of the DiT model. To mitigate the aforementioned domain gap, we coarsely voxelize the ground truth vertices during training to provide coarse shape priors in the RoPE embeddings. Ultimately, we found that a voxel resolution of 32 strikes a reasonable trade-off between having an artist-like topology and keeping shape details. We believe that improving shape information injection strategies and optimizing DiT training via optimal transport (OT) remains a promising direction for future research.

\begin{figure}[t!]
\centering
\includegraphics[width=\linewidth]{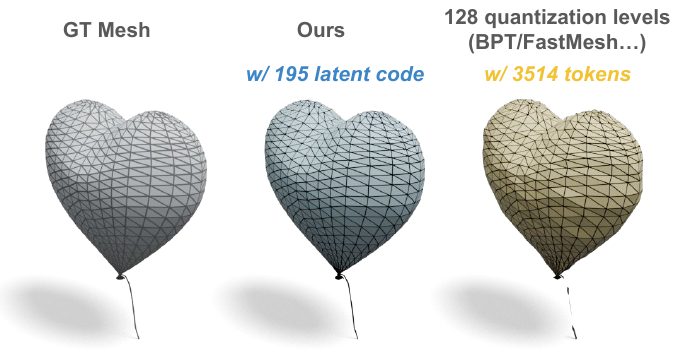}
\caption{
Qualitative comparisons with other mesh encoders.
AR-based methods necessitate mesh quantization.
Our MeshVAE works in continuous space, enabling the faithful preservation of the fine details in the input mesh.
}%
\label{fig:vae_comparisons}
\vspace{-5mm}
\end{figure}
\begin{figure}[t!]
\centering
\includegraphics[width=\linewidth]{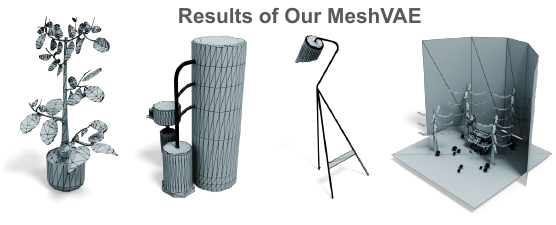}
\caption{
Reconstruction results of our MeshVAE\@.
Our MeshVAE faithfully retains the topological information of the input geometry within the continuous latent space.
Please refer to the supplementary for more results.
}%
\label{fig:vae_results}
\vspace{-5mm}
\end{figure}

\section{Experiments}%
\label{sec:exp}

We validate our method, \method, on a variety of different 3D shapes to study its generality.
We present qualitative and quantitative results in \cref{sec:meshvae_evaluation,sec:meshgen_evaluation}, respectively.
We compare \method against several competitors, demonstrating its effectiveness and efficiency in relation to those.
We also conduct ablation studies to validate the effectiveness of each component in our framework, as described in \cref{sec:ablation}.
Additional visualizations are given in our accompanying video and supplementary.

\begin{figure*}[t!]
\centering
\includegraphics[width=0.85\linewidth]{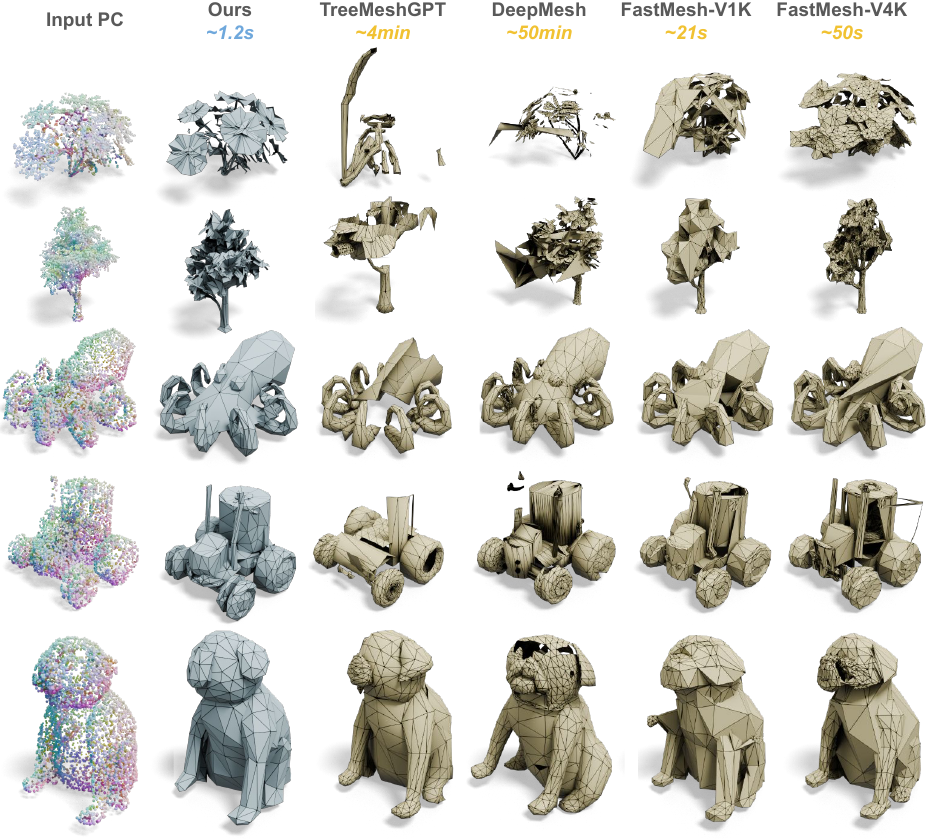}
\caption{
Qualitative comparisons with baseline methods for mesh generation conditioned on a point cloud.
The AR-based methods require significantly longer inference and frequently encounter early stopping, which often results in incomplete geometry.
In contrast, our diffusion-based method generates high-quality meshes efficiently.
}%
\label{fig:comparisons}
\end{figure*}

\subsection{Implementation Details}%
\label{sec:implementation}

We utilize an Objaverse-like~\cite{deitke23objaverse:} proprietary dataset of approximately $600k$ high-quality 3D models from professional artists.
MeshVAE utilizes an 8-layer Transformer backbone for the encoder and decoder.
It features a 1024-dimensional hidden size, resulting in a total of 233M parameters.
The DiT uses a standard transformer with 18 blocks and a 1024 hidden layer size, amounting to 427M parameters.
To speed up the DiT, we use Flash Attention and BF16 mixed precision during training, along with Exponential Moving Average (EMA) to improve stability and generalization.
See the supplementary for more details.

\subsection{Evaluation of MeshVAE}%
\label{sec:meshvae_evaluation}

\begin{table}[t]
\setlength{\tabcolsep}{3.0pt}
\caption{
Quantitative comparison of MeshVAE\@.
The Chamfer Distances (CD) are scaled by a factor of 100.
Although some methods share the same $128$-level quantization, failures on specific meshes can still shift the average score.
}%
\label{table:vae_comparision}
\centering
\begin{tabular}{clcc}
\toprule
Type                  & Method                                   & CD $\downarrow$  & Comp. Ratio$\downarrow$ \\ \midrule
AR                    & TreeMeshGPT~\cite{lionar2025treemeshgpt} & 1.63             & \underline{0.22}        \\
ArAE                  & EdgeRunner~\cite{tang2024edgerunner}     & \textbf{1.21}    & 0.47                    \\ \midrule
Diffusion             & Ours                                     & \underline{1.29} &\textbf{0.014}           \\ \bottomrule
\end{tabular}
\vspace{-3mm}
\end{table}

We first evaluate MeshVAE's ability to encode and decode meshes accurately.
We thus feed test meshes to the encoder, reconstruct them using the decoder, and compare the results to the original meshes using these metrics:
the Chamfer Distance (CD) between the reconstructed mesh and the GT mesh, which are normalized within $[-0.5, 0.5]$ and the Compression Ratio.
We use the \textit{original continuous mesh} as the ground truth, so the reported errors account for the losses due to quantization, for the methods that use it.
We report these results in \cref{table:vae_comparision}.

We compare MeshVAE to several alternative mesh encoders:
(1) MeshAnything~\cite{chen24meshanything}, which employs a na\"{\i}ve encoder, generating $9n_f$ tokens for a mesh with $n_f$ faces;
(2) BPT~\cite{weng25scaling}, which introduces block-wise indexing and patchified aggregation to achieve a $72\%$ token reduction;
(3) TreeMeshGPT~\cite{lionar2025treemeshgpt} and MeshSilkSong~\cite{song2025meshsilksong}, the current SOTA, which reduces token counts by $78\%$; and
(4) EdgeRunner~\cite{tang2024edgerunner}, which uses Auto-regressive Auto-encoder (ArAE) for mesh generation.
We could not evaluate MeshCraft~\cite{he25meshcraft:} directly, as the code is not yet available, however, since it quantizes the coordinates to $128$ values, we expect comparable performance to the above encoders.

\Cref{fig:vae_comparisons} presents a qualitative comparison of our proposed continuous-space reconstruction against other approaches.
\Cref{fig:vae_results} showcases additional high-fidelity reconstruction results.
Our results show that the representation learned by MeshVAE is efficient,  accurate due to the lack of quantization, and compact.

\subsection{Evaluation of Mesh Generation}%
\label{sec:meshgen_evaluation}

To quantitatively evaluate the generation conditioned on point clouds, we follow FastMesh and utilize the standard Toys4K dataset~\cite{stojanov21cvpr}.
Crucially, none of the compared models were trained on this dataset, thus ensuring a fair test of generalization.
We compare our model against several state-of-the-art approaches for shape-conditioned artistic mesh generation: MeshAnything~\cite{chen24meshanything,chen2024meshanythingv2}, TreeMeshGPT~\cite{lionar2025treemeshgpt}, BPT~\cite{weng2024bpt}, and FastMesh~\cite{kim2025fastmesh}.
We normalize both the generated and ground-truth shapes and sample 5,000 points from the GT mesh.
We adopt two standard metrics: Chamfer Distance (CD), which captures the overall structural similarity by measuring the average closest point distance, and Hausdorff Distance (HD), which quantifies the maximum discrepancy and is therefore highly sensitive to local errors.
As shown in \cref{table:ditcomparision}, our method achieves the lowest CD and HD scores.
Moreover, it is significantly faster than all other, auto-regressive, methods, demonstrating the superiority of our Rectified Flow model.
We also present more complex comparative visual results in \cref{fig:comparisons}.
While auto-regressive methods often suffer from early stopping, which leads to the loss of mesh geometry and severely degrades visual quality, our approach evades this issue.

We also note a limitation of these standard metrics.
Since our method directly predicts vertices and normal directly, CD and HD are very good.
In fact,  we omit the Normal Consistency (NC) metric as proposed in~\cite{chen2024meshanythingv2} as it is abnormally high for our method, but that does not mean much given that we predict normals directly.
In general, these metrics are not indicative of all aspects of mesh quality, and especially of its topology.
For these, we refer the reader to the qualitative comparisons.

\begin{table}[t]
\setlength{\tabcolsep}{4pt}
\caption{Quantitative comparison of shape-conditioned mesh generation on the Toys4K dataset.
The CD and HD are scaled by a factor of 100.
To ensure consistency, we adopted the ``Inf. Time'' calculation used by FastMesh, which reports the average inference time of a batch of multiple objects.  We report the time consumed for sampling and mesh extraction.
Notably, processing a single object with AR-based methods often requires $6\times$ the reported time, whereas our method maintains a constant runtime.
}%
\label{table:ditcomparision}
\centering
\small
\begin{tabular}{lccccc}
\toprule
Method                                      & CD $\downarrow$ & HD $\downarrow$ & Inf. Time(s) $\downarrow$ & \#V    \\ %
\midrule
MeshAnything~\cite{chen2024meshanything}*   & 12.02           & 26.87           & 26.06                     & 218.6  \\
MeshAnythingV2~\cite{chen24meshanything}*   & 10.23           & 24.98           & 31.94                     & 533.3  \\
TreeMeshGPT~\cite{lionar2025treemeshgpt}*   & 5.46            & 13.96           & 27.32                     & 706.3  \\
BPT~\cite{weng2024bpt}*                     & 5.71            & 12.02           & 49.23                     & 525.5  \\
FastMesh-V1K~\cite{kim2025fastmesh}*        & 4.09            & 10.32           & 3.41                      & 467.2  \\ 
FastMesh-V4K~\cite{kim2025fastmesh}*        & 4.05            & 10.22           & 6.60                      & 1040.6 \\ 
\midrule
Ours                                        & \textbf{2.33}   & \textbf{4.23}   & \textbf{1.06 + 0.47}              & 459.75  \\
\bottomrule
\multicolumn{5}{r}{*values from FastMesh~\cite{kim2025fastmesh}} \\
\end{tabular}
\vspace{-3mm}
\end{table}

\subsection{Ablation Studies}%
\label{sec:ablation}

We evaluate design choices and conduct an ablation study,
proving the importance of each component in the generation of high-quality 3D meshes.
\cref{table:vae_comparision} contains the results.

\paragraph{Different Downsample/Upsample Designs.}

We conduct a comparative study on token downsampling strategies, specifically focusing on the initialization of query tokens during cross-attention.
We ablate the choice of three different approaches:
(1) a Q-Former-like method using randomly initialized learnable tokens as queries;
(2) a Shape2Vecset-like approach employing Farthest Point Sampling (FPS) on encoded vertex features;
(3) our method, proposed in \cref{sec:contrastive_vae_training}, which utilizes a TokenMerge and MLPs, similar to a pixel-shuffle operation~\cite{shi2016pixelshuffle,chen2024interlvl}, to generate the initial queries.
As demonstrated in \cref{table:vae_ablation}, the first two sampling-based methods severely impeded training convergence and drastically reduced mesh reconstruction quality.
TokenMerge is thus key to ensuring stable training and achieving high-fidelity mesh outputs.

\paragraph{Impact of Different Compression Ratios.}

We also investigated the influence of varying downsampling ratios on reconstruction performance.
The results were intuitive: employing a higher downsampling ratio led to a corresponding decrease in model reconstruction performance.
However, we found that even with a significant $4\times$ downsampling ratio, as shown in \cref{table:vae_ablation}, the model was still capable of recovering excellent global and local shape structures, achieving an edges F1 score of approximately $0.89$.

\paragraph{Impact of Different Numbers of Vertices.}

We also explored the effect of different numbers of vertices on VAE reconstruction quality.
We created three datasets with different maximum vertex counts by modifying various filtering conditions, and then conducted testing.
As shown in \cref{table:vae_ablation}, we found that the reconstruction quality of our proposed method did not significantly degrade as the number of vertices increased, which further demonstrates the good scalability of our proposed representation.

\begin{table}[t]
\setlength{\tabcolsep}{1.0pt}
\caption{Ablation studies of different MeshVAE settings.
All values are scaled by a factor of 100.
}%
\label{table:vae_ablation}
\centering
\begin{tabular}{lccc}
\toprule
Method                & Vert. Dist.$\downarrow$ & Normals Dist.$\downarrow$ & F1 Score$\uparrow$ \\
\midrule
Q-former              & 23.36                   & 18.77                     & 49.47              \\
FPS                   & 18.29                   & 14.61                     & 60.18              \\
TokenMerge            & 0.75                    & 0.47                      & 99.78              \\
\midrule
downsample $\times$ 4 & 1.25                    & 1.30                      & 88.82              \\
downsample $\times$ 2 & 0.97                    & 1.11                      & 92.65              \\
\midrule
2048 Vertices         & 0.82                    & 0.73                      & 99.84              \\
4096 Vertices         & 0.75                    & 0.47                      & 99.78              \\
8192 Vertices         & 0.94                    & 0.49                      & 99.71              \\
\bottomrule
\end{tabular}
\vspace{-3mm}
\end{table}

\section{Conclusion and Discussion}%
\label{sec:conclusion}

In this work, we introduced a framework for mesh generation that bypasses the efficiency and quality limitations of autoregressive models.
Our method leverages a compact, continuous MeshVAE to compress a complex mesh geometry into a low-dimensional latent space, paired with a Flow-based Diffusion Transformer for efficient generation.

Despite its effectiveness, our framework has limitations that warrant future work.
First, our method assumes triangular faces, while artists often prefer other polygons such as quads.
Second, we observe minor holes in some generated results due to inaccurate diffusion predictions requiring heuristic post-processing based on short-cycle detection; refining the diffusion process or adopting a more powerful DiT model may mitigate this.
Additionally, common metrics such as CD and HD struggle to effectively evaluate artifacts in the generated meshes, including flipped normals and holes.
Future work should focus on developing metrics that can evaluate mesh topology quality as well, though this remains challenging for generative tasks.

Finally, we do not consider the problem of texture generation.
Extending our model with UV mapping generation might facilitate the generation of high-quality textures and is a compelling direction for future work.

{
\small
\bibliographystyle{ieeenat_fullname}
\bibliography{vedaldi_general,vedaldi_specific,main}
}

\clearpage
\setcounter{page}{1}
\maketitlesupplementary
\renewcommand{\thesection}{\Alph{section}}
\newcommand{\embeddings}{\boldsymbol{H}}
\newcommand{\normals}{\boldsymbol{N}}

\section{Additional Implementation Details}%
\label{sec:more_implementation}
In this section, we describe the implementation in more detail, including the data preparation and training specifics.

We utilize a proprietary dataset of approximately $600k$ high-quality 3D models from professional artists.
All data are normalized to the $[-1, 1]$ unit space and pre-processed by merging duplicate and outlier vertices \emph{without} any quantization.
We excluded meshes with a maximum vertex degree exceeding $50$ from the dataset for all experiments.
Notably, the most complex model in our unfiltered dataset contains up to $15k$ faces.
The MeshVAE is built upon an 8-layer encoder and an 8-layer decoder Transformer architecture, with a hidden size of $1024$, resulting in a model with a total of $233$ million parameters.
We use $16$ Fourier encoding frequency bands for all features.
The hyperparameters are set as follows: the threshold is $\tau=0.6$, the negative sampling weight is $\lambda_{\text{neg}}=0.01$, and the KL divergence weight is $\lambda_{\text{kl}}=10^{-3}$ by default.
For training, we employed the AdamW optimizer with a learning rate of $5 \times 10^{-4}$, complemented by a warmup strategy and a cosine annealing schedule.
The VAE was trained with a batch size of 64 per GPU for 3 days using 32 H100 GPUs.

The DiT uses the same standard transformer with 16 blocks and a 1536 hidden size, amounting to 895M parameters. For dit training, we used a learning rate of $1 \times 10^{-4}$ and trained the model for 10 days on 64 H100 GPUs with a batch size of $32$ per GPU\@. We leverage Flash Attention~\cite{dao2023flashattention2} and BF16 mixed precision to accelerate training, complemented by Exponential Moving Average (EMA) to improve stability and generalization.

\begin{figure*}[h]
\centering
\includegraphics[width=\textwidth]{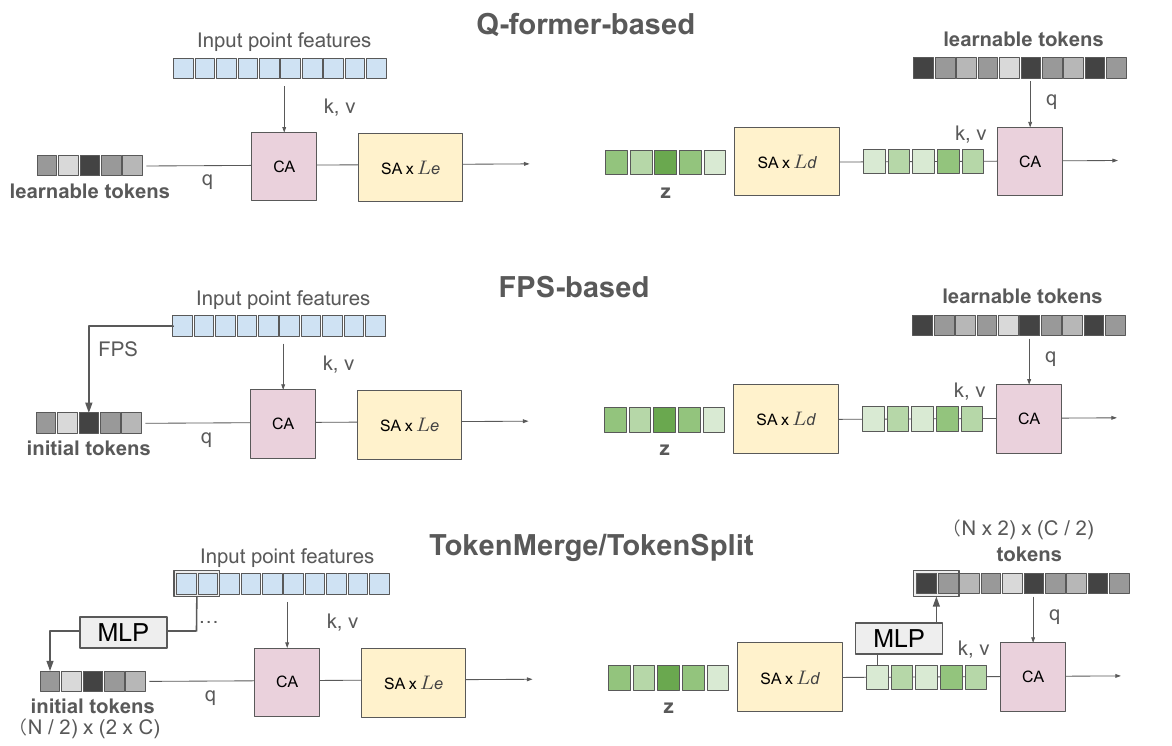}
\caption{
Detailed structure of different downsample and upsample strategies.
}%
\label{fig:downsample_upsample}
\vspace{-5mm}
\end{figure*}

\section{Details of the Mesh Extraction Algorithm}%
\label{sec:implementation-details}

\begin{algorithm}[t!]
\caption{Robust Mesh Extraction}%
\label{alg:mesh_extraction_algo}
\begin{algorithmic}[1]
\State \textbf{Input:} Mask logits $\boldsymbol{M} \in \mathbb{R}^{N}$, vertex coordinates $\vertices \in \mathbb{R}^{N \times 3}$, vertex normals $\normals \in \mathbb{R}^{N \times 3}$, edge embeddings $\embeddings \in \mathbb{R}^{N \times D}$ and edge distance threshold $\tau$
\State \textbf{Output:} Mesh $\mesh = (V, F)$

\Statex \textit{// Exact Edges and Adjacency Matrix}
\State $\mathbf{I} \gets \boldsymbol{M} > 0.5$
\State $\vertices \gets \vertices[\mathbf{I}]$, $\normals \gets \text{CLAMP}(\normals[\mathbf{I}], -1, 1)$
\State $D \gets \text{Compute Pairwise Distances}(\embeddings[\mathbf{I}])$
\State $E \gets \text{NonZero}((D > \tau) \land \text{UpperTriMask}(D))$
\State $\mathcal{A} \gets \text{Build Adjacency Matrix}(E)$

\Statex \textit{// Start Build All Faces from the Edges Soup}
\State $F_{\text{temp}} \gets \emptyset$
\For{$(v_1, v_2) \in \text{Unique}(E)$}
    \State $\mathcal{N}_{c} \gets (\mathcal{A}[v_1] \cap \mathcal{A}[v_2]) \setminus \{v_1, v_2\}$
    \State $F_{\text{temp}} \gets F_{\text{temp}} \cup \{\text{Sorted}(v_1, v_2, v_3) \mid v_3 \in \mathcal{N}_{c}\}$
\EndFor
\State $F \gets \text{Unique}(F_{\text{temp}})$

\Statex \textit{// Fix $k$-gon Ring in Topology}
\State $E_{\text{boundary}} \gets \text{Detect Boundary Edges}(E)$ 
\State $\mathcal{C}_{\text{loops}} \gets \text{Extract K-Loops}(\text{DFS}(E_{\text{boundary}}))$
\State $F \gets \text{Triangulate}(\mathcal{C}_{\text{loops}})$

\Statex \textit{// Recover Cyclic Ordering}
\State $N_{\text{face}} \gets \text{Normalize}(\text{Cross}(\vertex_{f,1} - \vertex_{f,0}, \vertex_{f,2} - \vertex_{f,0}))$ 
\State $N_{\text{avg}} \gets \text{Normalize}(\text{Mean}(\normal_F, \text{dim}=1))$
\State $F[N_{\text{face}} \cdot N_{\text{avg}} < 0, 1] \leftrightarrow F[N_{\text{face}} \cdot N_{\text{avg}} < 0, 2]$

\State \textbf{Return} $\text{Mesh}(V, F)$
\end{algorithmic}
\end{algorithm}

As shown in \cref{alg:mesh_extraction_algo}, we propose a simple and highly efficient mesh extraction algorithm to recover the local cyclic ordering of the vertices for each face and extract the final mesh from the decoded feature sequence $\hat{\mathbf{X}}$.
In sharp contrast to the method proposed by SpaceMesh~\cite{spacemesh2024}, which uses continuous permutation features to determine the local cyclic ordering of incident edges and infer the \emph{half-edge} connectivity, we leverage the additionally predicted vertex normal information to recover the edge ordering. 
Specifically, we first obtain a set of valid edges based on the edge embeddings.
Then, if an edge's two endpoints share a third vertex, the resulting triplet is considered to form a triangular face.
The cyclic ordering of the vertices within this face is determined by calculating the mean of the three vertex normals to serve as the face normal. 
Furthermore, to repair defects caused by inaccurate diffusion predictions, we implement a boundary repair post-processing step.
A boundary edge is defined as an edge that belongs to only one triangular face.
We therefore detect boundary edges and then determine whether these boundary edges form a $k$-gon ring.
By default, when $k < 5$, we perform triangulation to convert the loop into multiple triangular faces, thereby enhancing the robustness of the generated results.

\section{Different Downsample/Upsample Strategies}

We provide a detailed description of our Token Merge and Token Split operators for downsampling and upsampling, which are analogous to PixelShuffle.
We also detail the specific variations examined in the ablation studies.
\begin{itemize}
    \item A \textbf{Q-Former-based approach} that utilizes randomly initialized learnable tokens as queries for both downsampling and upsampling.
    \item A \textbf{FPS-based approach} that employs Farthest Point Sampling (FPS) on encoded vertex features during downsampling like Shape2Vecset~\cite{zhang2023shape2vecset}, and uses learnable tokens for upsampling.
    \item Our \textbf{TokenMerge/Split approach} that groups the encoded vertex features for downsampling and splits the latents by channel for upsampling like PixelShuffle.
\end{itemize}

\begin{figure*}[h]
\centering
\includegraphics[width=\textwidth]{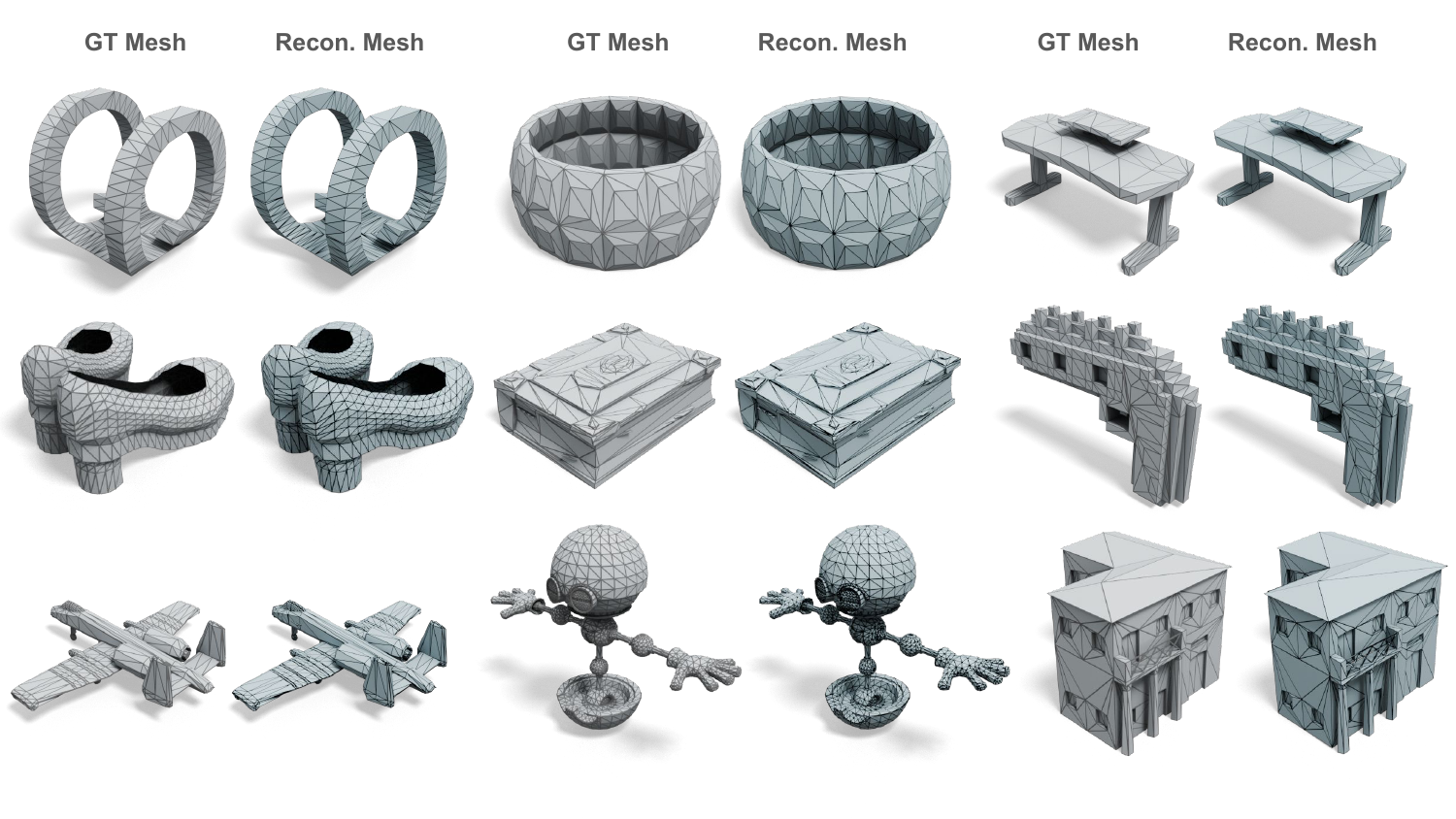}
\vspace{-15mm}
\caption{
More mesh reconstruction results of our MeshVAE.
}%
\label{fig:vae_recon}
\end{figure*}

\section{Inference Time of A Single Object}

As noted in Table 2 in the paper, to ensure consistency with FastMesh~\cite{kim2025fastmesh}, we initially adopted the same calculation method for ``Inf. Time'', which uses the average inference time of a batch of multiple objects.
However, this metric may not accurately reflect the true speed of single-object inference.
For a fair comparison, we present the time required for different methods to infer a single mesh.
The results in \cref{table:dittime} demonstrate the superior inference efficiency of our proposed diffusion-based method compared to auto-regressive methods.

\begin{figure}[h]
\centering
\includegraphics[width=\linewidth]{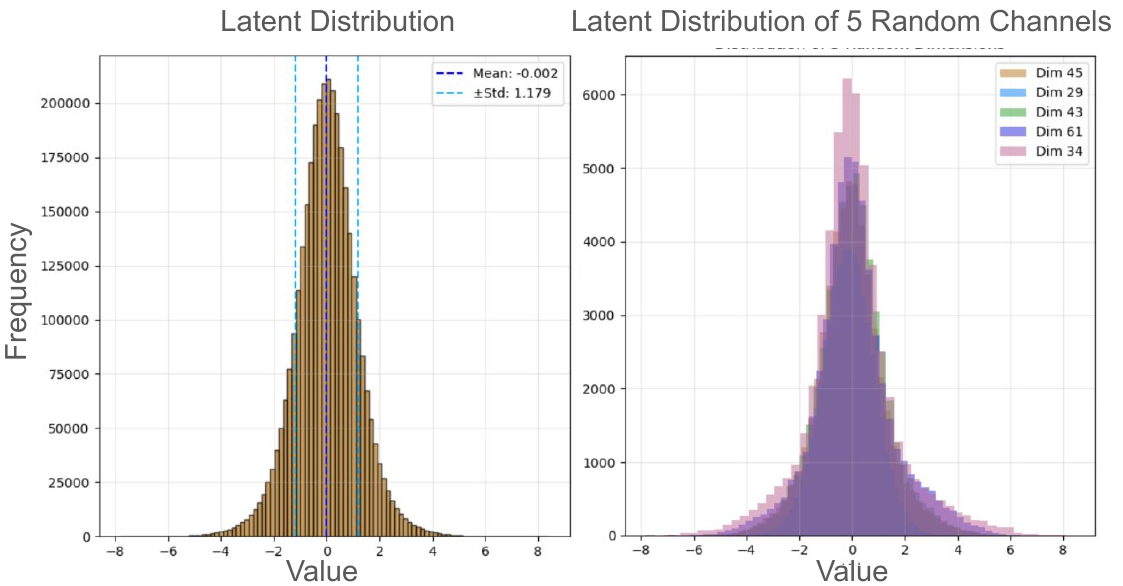}
\caption{
Latent distribution of the learned latent space.
}%
\label{fig:latent_distribution}
\vspace{-5mm}
\end{figure}

\begin{table*}[t]
\centering
\small
\caption{Inference time of a single mesh generation.
For autoregressive-based methods, the time varies with the complexity of the input.}%
\label{table:dittime}
\begin{tabular}{lcccccc}
\toprule
\textbf{Method} & \textbf{BPT~\cite{weng2024bpt}} & \textbf{TreeMeshGPT~\cite{lionar2025treemeshgpt}} & \textbf{DeepMesh~\cite{zhao2025deepmesh}} & \textbf{FastMesh-V1K~\cite{kim2025fastmesh}} & \textbf{FastMesh-V4K~\cite{kim2025fastmesh}} & \textbf{Ours} \\
\midrule
Inference Time & $\sim$8 min & $\sim$4 min & $\sim$50 min & $\sim$21 s & $\sim$50 s & $\sim$1.2 s \\
\bottomrule
\end{tabular}
\end{table*}

\section{Latent Space Learned by MeshVAE}

We compress the discrete mesh data structure into a continuous and compact latent space.
Here, we illustrate the distribution of the learned latent space of our MeshVAE, along with the latent distribution of a randomly sampled channel.
As shown in \cref{fig:latent_distribution}, the latent distribution closely approximates a standard Gaussian distribution.

\section{Additional Results}
\begin{figure}[h]
\centering
\includegraphics[width=\linewidth]{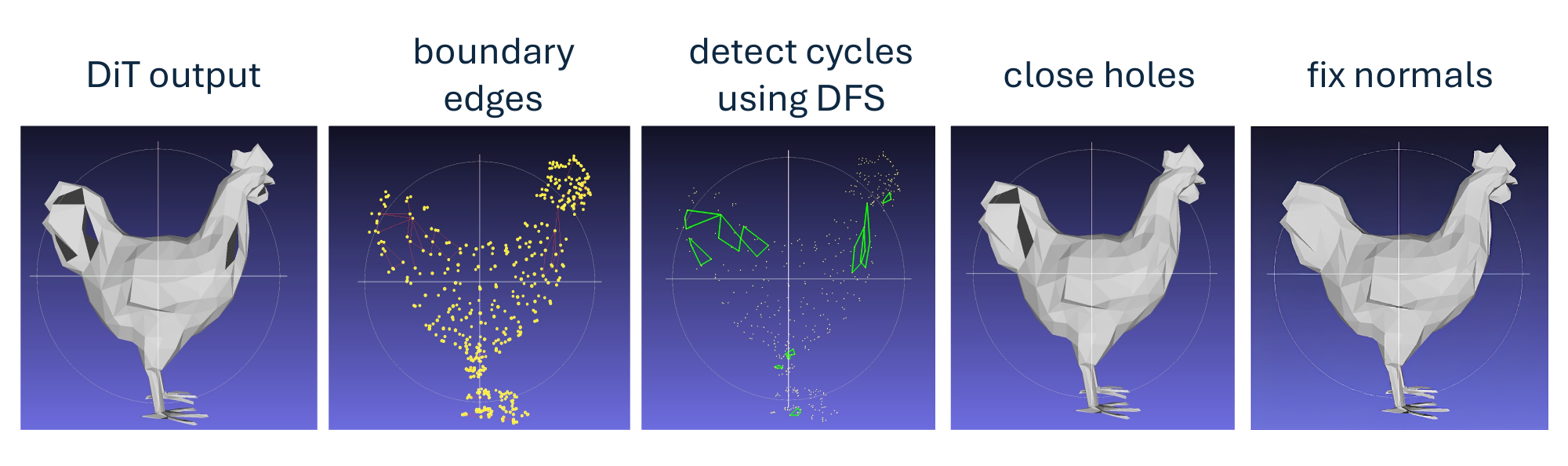}
\caption{
Post processing of our generated meshes.
}%
\label{fig:post_processing}
\vspace{-5mm}
\end{figure}
\begin{figure*}[h]
\centering
\includegraphics[width=0.88\textwidth]{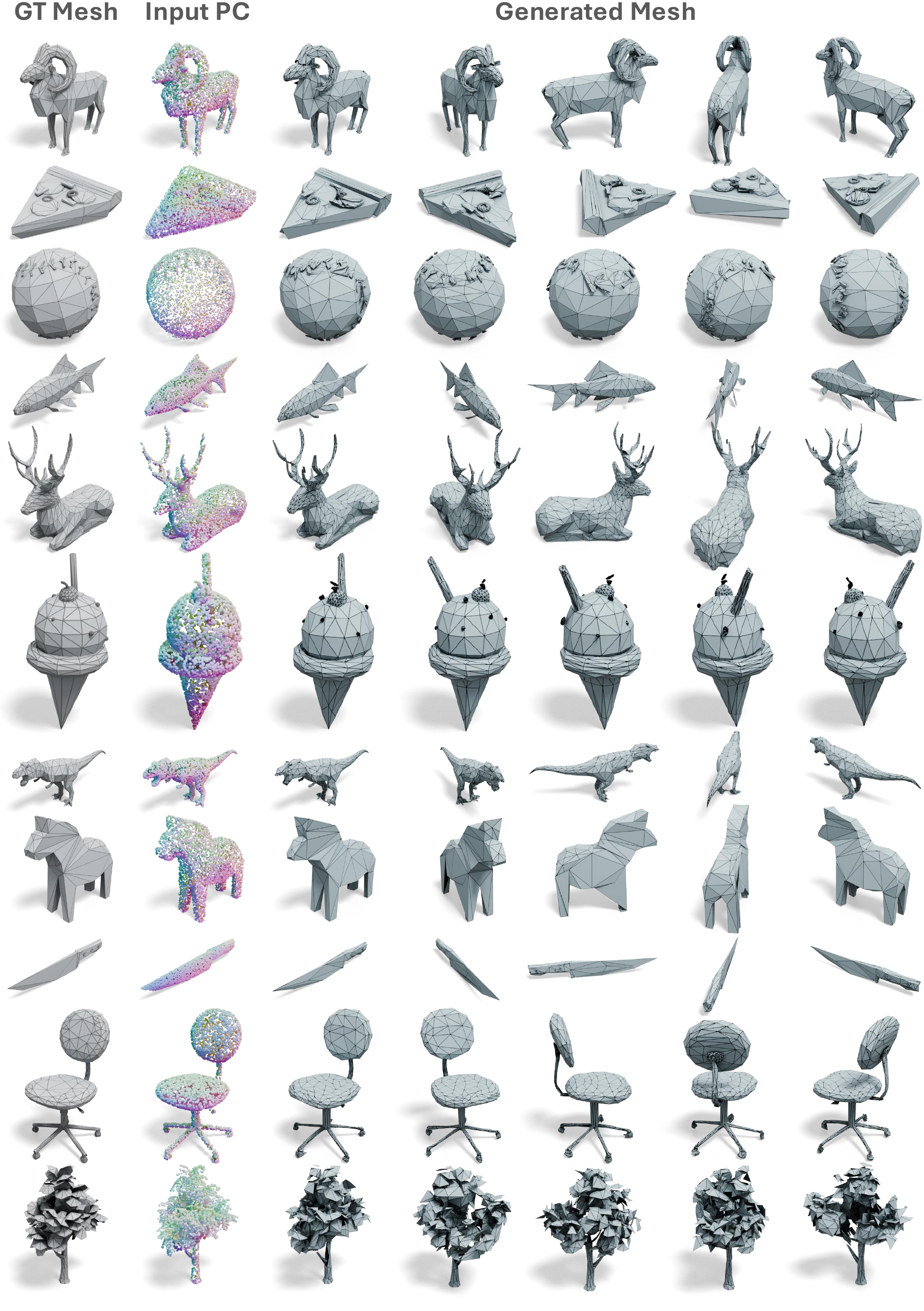}
\caption{
Additional point cloud conditioned mesh generation results for our method.
}%
\label{fig:more_dit}
\vspace{-5mm}
\end{figure*}

We present additional mesh reconstruction results of the MeshVAE in~\cref{fig:vae_recon}.
The results collectively demonstrate that our method is capable of high-fidelity reconstruction of the input mesh from the learned latent space, preserving critical geometric details.
Furthermore, we show more results of point cloud conditioned mesh generation i ~\cref{fig:more_dit}.
These results highlight the ability of our method to generate a diverse set of meshes that are faithful to the input point cloud guidance.
We also illustrated the detailed post-processing workflow in Fig.~\ref{fig:post_processing}.

\begin{figure}[h]
\centering
\includegraphics[width=\linewidth]{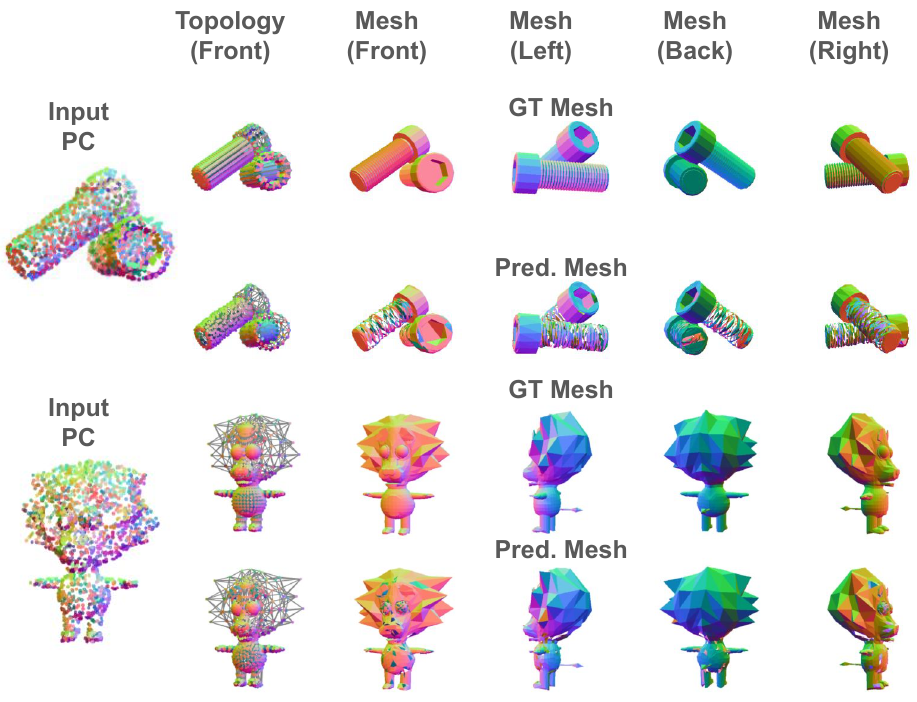}
\caption{
Failure cases of our generated meshes.
}%
\label{fig:failure_cases}
\vspace{-5mm}
\end{figure}

\section{Failure Cases}

Our method is also subject to some failure cases when the model has difficulty generating accurate latents during the denoising process, which in turn leads to noisy vertex positions and edge embeddings.
This ultimately causes the extracted meshes to exhibit holes, as shown in~\cref{fig:failure_cases}.
We believe that this could be further mitigated by carefully refining the diffusion process or adopting a more powerful DiT model for generation.

\end{document}